\documentclass[11pt,a4paper]{article}
\usepackage[hyperref]{acl2018}
\usepackage{times}
\usepackage{latexsym}

\usepackage{xcolor}
\usepackage{graphicx}
\usepackage{fancyvrb}
\usepackage{url}
\usepackage{paralist}
\usepackage{amsmath,amsfonts,amssymb}
\usepackage{booktabs}
\usepackage{multirow}

\usepackage{eqnarray}

\usepackage{url}

\aclfinalcopy 
\def\aclpaperid{448} 

\title{Personalizing Dialogue Agents: I have a dog, do you have pets too?}

\author{Saizheng Zhang$^{\dagger,1}$, Emily Dinan$^{\ddagger}$, Jack Urbanek$^{\ddagger}$, Arthur Szlam$^{\ddagger}$, Douwe Kiela$^{\ddagger}$, Jason Weston$^{\ddagger}$\\
   $^{\dagger}$ Montreal Institute for Learning Algorithms, MILA\\$^{\ddagger}$ Facebook AI Research\\
  {\tt saizheng.zhang@umontreal.ca, \{edinan,jju,aszlam,dkiela,jase\}@fb.com}}

\date{}

\begin{document}
\maketitle
\addtocounter{footnote}{1}
\footnotetext{Work done while at Facebook AI Research.}
\begin{abstract}
Chit-chat models are known to have several problems: they lack specificity, do not display a consistent personality and are often not very captivating.
In this work we present the task of
 making chit-chat more engaging by conditioning on 
 profile information.
We collect data and train models to (i) condition on
their given profile information; and (ii) information about the person they are talking to,
resulting in improved dialogues, as measured by next utterance prediction.
Since (ii) is initially unknown,
our model is trained to engage its partner with personal topics, and we show the resulting dialogue can be used to predict profile information about the interlocutors.
\end{abstract}

\newif\ifarxiv
\arxivfalse

\newif\ifnoarxiv
\noarxivtrue

\section{Introduction}

Despite much recent success in natural language processing and dialogue research, communication between a human and a machine is still in its infancy.
It is only recently that neural models have had sufficient capacity and access to sufficiently large datasets that they appear to generate meaningful responses in a chit-chat setting. Still, conversing with such generic chit-chat models for even a short amount of 
time quickly exposes their weaknesses \citep{serban2016generative,vinyals2015neural}.

Common issues with chit-chat models  
include:
(i) the lack of a consistent personality \citep{li2016persona} as they are typically trained over many dialogs each with different speakers,  (ii)
the lack of an explicit long-term memory as they are typically trained to produce an utterance given only the recent dialogue history \citep{vinyals2015neural}; and  (iii)
a tendency to produce non-specific answers like ``I don't know'' \citep{li2015diversity}. 
Those three problems combine to produce an unsatisfying overall experience for a human to engage with. We believe some of those problems are due to there being no good publicly available dataset for general chit-chat. 
\ifarxiv
\footnote{For example,  currently the  most general chit-chat dataset available in \url{http://parl.ai} a large repository of dialogue datasets is probably OpenSubtitles, which is based on movie scripts, not natural conversations.}.
\fi

Because of the low quality of current conversational models, and because of the difficulty in evaluating these models, chit-chat
is often ignored as an end-application.  Instead, the research community has focused on 
 task-oriented communication,
 such as airline or restaurant booking \citep{bordes2016learning}, or else single-turn information seeking, i.e. question answering \cite{rajpurkar2016squad}. 
Despite the success of the latter, simpler, domain,
it is well-known that a large quantity of human dialogue centers on socialization, personal interests and chit-chat \citep{dunbar1997human}. For example, less than 5\% of posts on Twitter are questions, whereas around 80\% are about personal emotional state, thoughts or activities, authored by so called ``Meformers'' \citep{naaman2010really}.

In this work we make a step towards more engaging chit-chat dialogue agents by endowing them with a configurable, but persistent persona, encoded by multiple sentences of textual description, termed a profile. This profile can be stored in a memory-augmented neural network and then used to produce more personal, specific, consistent and engaging responses than a persona-free model, thus alleviating some of the common issues in chit-chat models.
Using the same mechanism, any existing information about the persona of the dialogue partner can also be used in the same way. Our models are thus trained to both ask and answer questions about personal topics, and the resulting dialogue can be used to build a model of the persona of the speaking partner.

To support the training of such models, we present the {\sc persona-chat} dataset, a new dialogue dataset consisting of 162,064 utterances 
between crowdworkers who were randomly paired and each asked to act the part of a given provided persona (randomly assigned, and created by another set of crowdworkers). The paired workers were asked to chat naturally and to get to know each other during the conversation. This produces interesting and engaging conversations that our agents can try to learn to mimic. 


Studying the next utterance prediction task during dialogue, we compare a range of models: both generative and ranking models, including Seq2Seq models and Memory Networks \citep{memn2n} as well as other standard retrieval baselines. 
We show experimentally that in either the generative or ranking case 
conditioning the agent with persona information  
gives improved prediction of the next dialogue utterance.  
The {\sc persona-chat} dataset is designed to facilitate research into alleviating some of the issues that traditional chit-chat models face, and with the aim of making such models more consistent and engaging, by endowing them with a persona.
By comparing against chit-chat models built using the OpenSubtitles and Twitter datasets,
human evaluations show that our dataset provides more engaging models,
that are simultaneously capable of being fluent and consistent via conditioning on a persistent, recognizable profile.

\section{Related Work}

Traditional dialogue systems consist of building blocks, such as dialogue state tracking components and response generators, and have typically been applied to tasks with labeled internal dialogue state  and precisely defined user intent (i.e., goal-oriented dialogue), see e.g. \citep{young2000probabilistic}. 
The most successful goal-oriented dialogue systems model conversation as partially observable Markov decision processes (POMDPs) \citep{young2013pomdp}.
All those methods typically do not consider the chit-chat setting and are more concerned with achieving functional goals (e.g. booking an airline flight) than displaying a personality.
In particular, many of the tasks and datasets available are constrained to narrow domains \citep{serban2015survey}.

Non-goal driven dialogue systems go back to Weizenbaum's famous program ELIZA 
\citep{weizenbaum1966eliza}, and hand-coded systems have continued to be used in applications 
to this day. For example, modern solutions that build an open-ended dialogue system to the Alexa challenge  combine hand-coded and machine-learned elements \citep{serban2017deep}.
Amongst the simplest of statistical systems that can be used in this domain, that are based on data rather than hand-coding, are information retrieval models \citep{sordoni2015neural}, which retrieve and rank responses based on their matching score with the recent dialogue history.
We use IR systems as a baseline in this work.

End-to-end neural approaches are a class of models which have seen growing recent interest.  
A popular class of methods are
generative recurrent systems like seq2seq applied to dialogue \citep{sutskever2014sequence,vinyals2015neural,sordoni2015neural,li2016deep,serban2017hierarchical}.
\ifarxiv
Their strengths are that (i) they are not constrained by hard-code rules or explicit internal states that may work well in a narrow domain, but are too restrictive for more open dialogue such as chit-chat, and (ii) being based on architectures rooted in language modeling and machine translation, they excel at generating syntactically coherent language, and can generate entirely novel responses.
Their deficiencies are that they typically need a large amount of data to be trained, 
and the vanilla approach generates responses given only the recent dialogue history without using other external memory. 
The latter issue makes neural models  hence typically lack both domain knowledge in the domain being discussed, and a persistent personality during discussions.
\else
Rooted in language modeling, they are able to produce syntactically coherent novel responses, but
their memory-free approach means they lack long-term coherence and a persistent personality, as discussed before. 
\fi
A promising direction, that is still in its infancy, to fix this issue is to use 
a memory-augmented network instead \citep{memn2n,dodge2015evaluating} 
by providing or learning appropriate  memories.
\ifarxiv
A related class of neural methods is to retrieve and rank candidates
rather than generate words,
 similarly to IR methods, but using memory-augmented networks to score the candidates instead. We compare the generative and ranking approaches to each other in this work.
\fi

\newcite{serban2015survey} list available corpora for training dialogue systems.
Perhaps the most relevant to learning chit-chat models are ones based on movie
scripts such as OpenSubtitles and Cornell Movie-Dialogue Corpus, and dialogue from web platforms such as Reddit and Twitter, all of which have  been used for training neural
approaches \citep{vinyals2015neural,dodge2015evaluating,li2016deep,serban2017hierarchical}. Naively training on these datasets leads to models with the lack of a consistent personality as they will learn a model averaged over many different speakers. Moreover, the data does little to encourage the model to engage in understanding and maintaining knowledge of the dialogue partner's personality and topic interests.

According to \newcite{serban2015survey}'s survey, personalization of dialogue systems is ``an
important task, which so far has not received much attention''. In the case of goal-oriented dialogue some work has focused
on the agent being aware of the human's profile and adjusting the dialogue accordingly, but without a personality to the agent itself
 \citep{lucas2009managing,joshi2017personalization}.
For the chit-chat setting, the most relevant work is \citep{li2016persona}.
For each user in the Twitter corpus, personas were captured via distributed embeddings (one per speaker) to encapsulate individual characteristics such as background information and speaking style,
and they then showed using those vectors improved the output of their seq2seq model for the same speaker. 
Their work does not focus on attempting to engage the other speaker by getting to know them, as we do here. For that reason,
our focus is on explicit profile information, not hard-to-interpret latent variables.

\begin{table*}[t]
  \begin{center}
    \begin{small}
      \begin{tabular}{l|l}
        \toprule
        \textbf{Original Persona} & \textbf{Revised Persona}\\
        \midrule
I love the beach. & To me, there is nothing like a day at the seashore. \\
My dad has a car dealership & My father sales vehicles for a living. \\
I just got my nails done & I love to pamper myself on a regular basis. \\
I am on a diet now & I need to lose weight. \\
Horses are my favorite animal. &  I am into equestrian sports. \\
\midrule
\ifarxiv
I am an eccentric hair stylist for dogs & I work with animals. \\ 
My favorite past time is collecting Civil War antiques. & I like finding or buying historical artifacts. \\
I fake a British accent to seem more attractive. & I heard girls liked foreigners.\\
I have been married four times and widowed three. & I have a lot of experience with marriage\\
I have an allergy to mangoes & I have reactions to certain fruits. \\
\midrule
\fi
I play a lot of fantasy videogames. & RPGs are my favorite genre. \\
I have a computer science degree. & I also went to school to work with technology. \\
My mother is a medical doctor & The woman who gave birth to me is a physician. \\
I am very shy. & I am not a social person.\\
I like to build model spaceships.& I enjoy working with my hands. \\
\bottomrule
      \end{tabular}
      \caption{Example Personas (left) and their revised versions (right) from the {\sc persona-chat} dataset.
The revised versions are designed to be characteristics that the same persona might have, which could be rephrases, 
generalizations or specializations.
 \label{table:persona-examples}}
    \end{small}
  \end{center}
\end{table*}

\begin{table*}[t]
  \begin{center}
    \begin{small}
      \begin{tabular}{l|l}
        \toprule
        \textbf{Persona 1} & \textbf{Persona 2}\\
        \midrule
I like to ski & I am an artist\\
My wife does not like me anymore & I have four children\\
I have went to Mexico 4 times this year & I recently got a cat \\
I hate Mexican food &  I enjoy walking for exercise \\
I like to eat cheetos &  I love watching Game of Thrones\\
\bottomrule
\multicolumn{2}{l}{ }\\
\multicolumn{2}{l}{[PERSON 1:] Hi}\\
\multicolumn{2}{l}{[PERSON 2:] Hello ! How are you today ?}\\
\multicolumn{2}{l}{[PERSON 1:] I am good thank you , how are you.}\\
\multicolumn{2}{l}{[PERSON 2:] Great, thanks ! My children and I were just about to watch Game of Thrones. }\\
\multicolumn{2}{l}{[PERSON 1:] Nice ! How old are your children?}\\
\multicolumn{2}{l}{[PERSON 2:] I have four that range in age from 10 to 21. You?}\\
\multicolumn{2}{l}{[PERSON 1:] I do not have children at the moment.}\\ 
\multicolumn{2}{l}{[PERSON 2:] That just means you get to keep all the popcorn for yourself.}\\
\multicolumn{2}{l}{[PERSON 1:] And Cheetos at the moment!}\\
\multicolumn{2}{l}{[PERSON 2:] Good choice. Do you watch Game of Thrones?}\\
\multicolumn{2}{l}{[PERSON 1:] No, I do not have much time for TV.}\\
\multicolumn{2}{l}{[PERSON 2:] I usually spend my time painting: but, I love the show.}\\
      \end{tabular}
      \caption{Example dialog from the {\sc persona-chat} dataset. Person 1 is given their own persona (top left) at the beginning of the chat, but does not know the persona of Person 2, and vice-versa. They have to get to know each other during the conversation.
 \label{table:persona-chat-example}}
    \end{small}
  \end{center}
\end{table*}

\section{The {\sc persona-chat} Dataset} 

The aim of this work is to facilitate more engaging and more personal chit-chat dialogue. The {\sc persona-chat} dataset is a crowd-sourced dataset, collected via Amazon Mechanical Turk, where each of the pair of speakers condition their dialogue on a given profile, which is provided. 

The data collection consists of three stages:


(i) Personas: we crowdsource a set of 1155 possible personas, each consisting of at least 5 profile sentences, setting aside 100 never seen before personas for validation, and 100 for test.

(ii) Revised personas: to avoid modeling that takes advantage of trivial word overlap, we crowdsource  additional rewritten sets of the same 1155 personas, with related sentences that are rephrases, generalizations or specializations, rendering the task much more challenging.

(iii) Persona chat: we pair two Turkers and assign them each a random (original) persona from the pool, and ask them to chat. This resulted in a dataset of 162,064 utterances over 10,907 dialogs, 15,602 utterances (1000 dialogs) of which are set aside for validation, and 15,024 utterances
(968 dialogs) for test.


The final dataset and its corresponding data collection source  code, as well as models trained on the data, are all available open source in ParlAI\footnote{ {\small{\url{http://parl.ai}}}}.

In the following, we describe each data collection stage and the resulting tasks in more detail.

\subsection{Personas}

We asked the crowdsourced workers to create a character (persona) description using 5 sentences, providing them only a single example:

{\em ``I am a vegetarian. I like swimming.  My father used to work for Ford.  My favorite band is Maroon5. I got a new job last month, which is about advertising design.''}

Our aim was to create profiles that are natural and descriptive, and contain typical topics of human interest that the speaker 
can bring up in conversation. Because the personas are not the real profiles of the Turkers,
  the dataset does not contain personal information (and they are told specifically not to use any).
We asked the workers to make each sentence short, with a maximum of 15 words per sentence.
This is advantageous both for humans and machines: if they are too long, crowdsourced workers are likely to lose interest, and for machines the task could become more difficult.

Some examples of the  personas collected are given in Table \ref{table:persona-examples} (left).

\subsection{Revised Personas}

A difficulty when constructing dialogue datasets, or text datasets in general, is that in order to
encourage research progress, the task must be carefully constructed so that is neither too
 easy nor too difficult for the current technology 
 \citep{voorhees1999trec}.
One issue with conditioning on textual personas is that there is a danger that humans will, even if asked not to,
unwittingly repeat profile information either verbatim or with significant word overlap.
This may make any subsequent machine learning tasks less challenging, and the solutions will not generalize to
more difficult tasks. This has been a problem in some recent datasets:
for example, the dataset curation technique used for the well-known SQuAD dataset
suffers from this word overlap problem to a certain extent \citep{chen2017reading}.

To alleviate this problem, we presented the original personas we collected to a new set of crowdworkers
and asked them to rewrite the sentences so that a new sentence is about 
{\em ``a related characteristic that the same person may have''},
hence the revisions could be rephrases, generalizations or specializations.
For example {\em ``I like basketball''} can be revised as {\em ``I am a big fan of Michael Jordan''}
not because they mean the same thing but because the same persona could contain both. 

In the revision task, workers are instructed not to trivially rephrase the sentence by copying the original words.
However, during the entry stage if a non-stop word is copied we issue a
warning, and ask them to rephrase, guaranteeing that the instructions are followed. 
For example, {\em ``My father worked for Ford.''} can be revised to
{\em ``My dad worked in the car industry''}, but not
{\em ``My dad was employed by Ford.''} due to word overlap.

\ifarxiv
Finally, we encourage the construction of  natural sentences. In earlier versions of the task we noticed that
the word overlap constraint caused unwanted unnatural constructions such as {\em ``I like eating pretzels''} revised as
{\em ``I like to chew and swallow twisted bread with salt''}. Giving explicit instructions about this seemed to help,
where we prefer a revision like {\em ``I enjoy beers and beer snacks''}.
\fi

Some examples of the revised personas collected are given in Table \ref{table:persona-examples} (right).

\subsection{Persona Chat}\label{personachatter}

After collecting personas, we then collected the dialogues themselves, conditioned on the personas.
For each dialogue, we paired two random crowdworkers, and gave them the instruction that they will chit-chat with another worker, while
playing the part of a given character. We then provide them with a randomly chosen persona from our pool, different to their partners.
The instructions are on purpose quite terse and simply ask them to 
``chat with the other person naturally and try to get to know each other''.
In an early study we noticed the crowdworkers tending to talk about themselves (their own persona) too much, so
we also added the instructions
``both ask questions and answer questions of your chat partner'' which seemed to help.
We also gave a bonus for high quality dialogs.
The dialog is turn-based, with a maximum of 15 words per message.
We again gave instructions to not  trivially copy the character descriptions into the messages,
but also wrote explicit code sending them an error if they tried to do so, using simple string matching.
We define a minimum dialogue length which is randomly between 6 and 8 turns each for each dialogue.
An example dialogue from the dataset is given in Table  \ref{table:persona-chat-example}.

\subsection{Evaluation}

We focus on the standard dialogue task of predicting the next utterance given the dialogue history, but consider this task both with and without the profile information being given to the learning agent. Our goal is to enable interesting directions for future research, where chatbots can for instance have personalities, or imputed personas could be used to make dialogue more engaging to the user.

We consider this in four possible scenarios: conditioning on no persona, your own persona, their persona, or both. These scenarios can be tried using either the original personas, or the revised ones.
We then evaluate the task using three metrics: (i) the log likelihood of the correct sequence, measured via perplexity, (ii) F1 score, and (iii) 
next utterance classification loss, following \newcite{lowe2015ubuntu}.
The latter consists of choosing $N$ random distractor responses from other dialogues (in our setting, $N$=19) and the model selecting the best response among them, resulting in a score of one if the model chooses the correct response, and zero otherwise (called hits@1 in the experiments). 

\ifarxiv
\begin{figure*}[t]
	\centering
	\includegraphics[width=\textwidth]{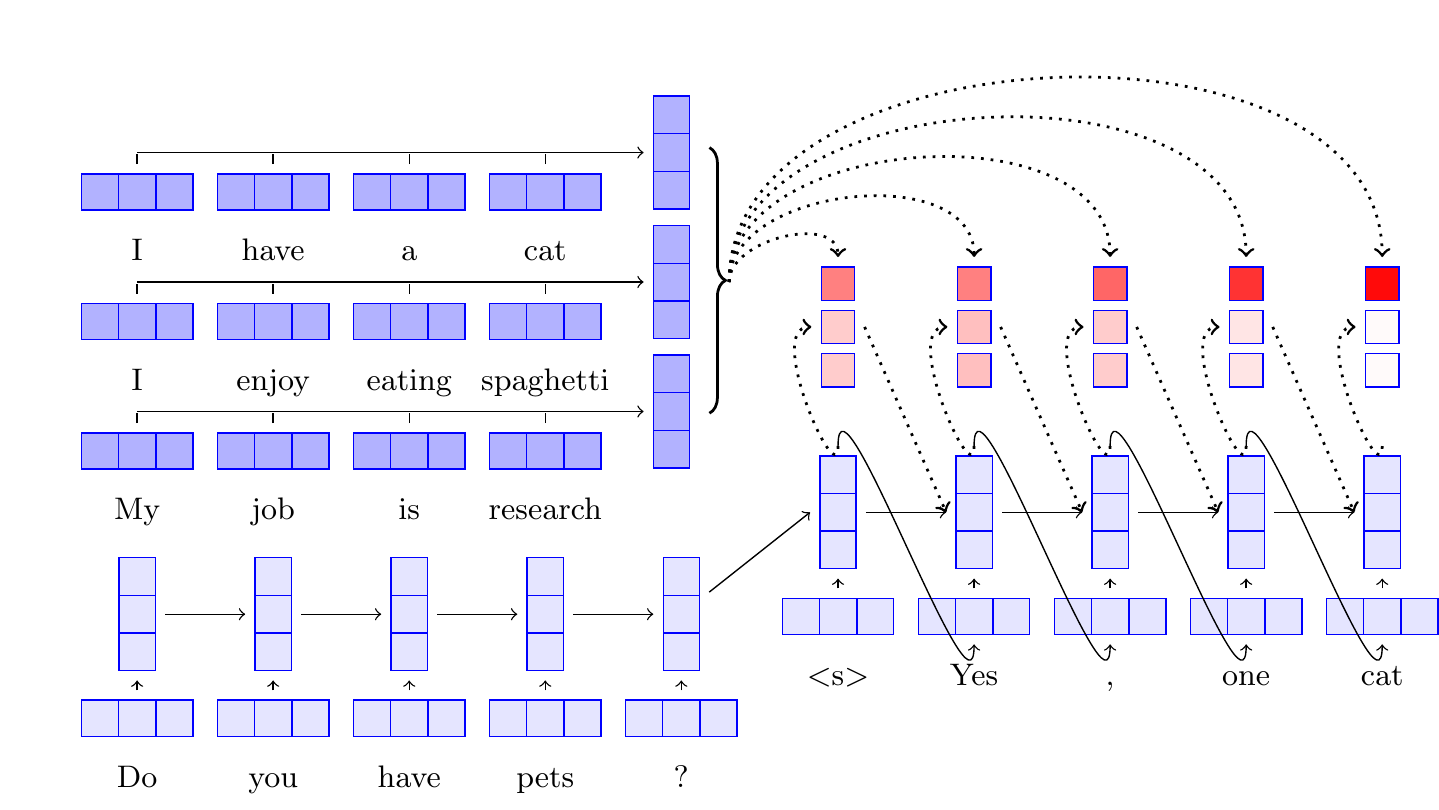}
    \caption{\label{fig:PMN-gen}A diagram of the Profile Memory Network for generation. We also implemented a ranking version which has the same architecture except it ranks candidate sentences from the training set instead of generating, representing them using bag-of-word embeddings.}
\end{figure*}
\fi

\section{Models}

We consider two classes of model for next utterance prediction: ranking models and generative models.
Ranking models produce a next utterance by considering any utterance in the training set as a possible candidate reply. Generative models 
 generate novel sentences by conditioning on the dialogue history (and possibly, the persona), and then generating the response word-by-word. 
Note one can still evaluate the latter as ranking models by computing the probability of generating a given candidate, and ranking candidates by those scores.

%


\ifarxiv
\subsubsection{IR Baseline}
To select candidate responses a standard baseline is nearest neighbor information retrieval (IR) (Isbell et al., 2000; Jafarpour et al., 2010; Ritter et al., 2011; Sordoni et al., 2015). 
While there are many variants, we adopt the simplest one: find the most similar message in the (training) dataset and output the response from that exchange. Similarity is measured by the tf-idf weighted cosine similarity between the bags of words. To incorporate the profile we simply concatenate it to the query vector bag of words.

\subsubsection{Starspace}

Starspace is a recent model that performs also performs information retrieval but by learning sentence embeddings that measure similarity between the dialog and the next utterance by optimizing the word embeddings directly for that task using the training set \citep{wu2017starspace}\footnote{Available at \url{https://github.com/facebookresearch/StarSpace}}. Similar supervised embeddings have been used with good results in other dialogue tasks previously \citep{dodge2015evaluating}. 
Specifically, it optimizes:
\[
\label{math:loss}
\sum_{\substack{(q,c) \in E^+\\ b^{-} \in E^{-}}} L(sim(q, c), sim(q, c^{-}_1), \dots, sim(q, c^{-}_k))
\]
where the loss function $L$ compares a positive pair of query and candidate $(q,c)$ with the negative pairs $(q, c^{-}_i)$, $i=1,\dots, k$ using the margin ranking loss $\max(0,\mu - sim(q,c)$, where $\mu$ is the margin parameter. The similarity function $sim(\cdot, \cdot)$ is the cosine similarity of the sum of word embeddings of the query $q$ and candidate $c'$. Denoting the dictionary of ${\cal D}$ word embeddings as $W$ which is a ${\cal D} \times d$ matrix, where $W_i$ indexes the $i^{th}$ word (row), yielding its $d$-dimensional embedding, it embeds inputs $q$ and $c'$. 
  with $\sum_{i \in s} W_i$. While this model supports different word embeddings for the left and right hand side of the similarity function, we found sharing the weights gave the best performance.
  
Similar to the IR baseline, to incorporate the profile we simply concatenate it to the query vector bag of words. 
\else
\subsection{Baseline ranking models}
We first consider two baseline models, 
an IR baseline \cite{sordoni2015neural} and a supervised embedding model, Starspace \cite{wu2017starspace}\footnote{\url{github.com/facebookresearch/StarSpace}}. While there are many IR variants, we adopt the simplest one: find the most similar message in the (training) dataset and output the response from that exchange. Similarity is measured by the tf-idf weighted cosine similarity between the bags of words. 
Starspace is a recent model that also performs information retrieval but by learning 
the similarity
between the dialog and the next utterance by optimizing the embeddings directly for that task using the margin ranking loss and $k$-negative sampling. The similarity function $sim(q, c')$ is the cosine similarity of the sum of word embeddings of the query $q$ and candidate $c'$. Denoting the dictionary of ${\cal D}$ word embeddings as $W$ which is a ${\cal D} \times d$ matrix, where $W_i$ indexes the $i^{th}$ word (row), yielding its $d$-dimensional embedding, it embeds the sequences $q$ and $c'$.

In both methods, IR and StarSpace, to incorporate the profile we simply concatenate it to the query vector bag of words. 
\fi

\subsection{Ranking Profile Memory Network}

Both the previous models use the profile information by combining it 
with the dialogue history, which means those models cannot differentiate between the two when deciding on the next utterance. In this model we instead use a memory network with the dialogue history as input, which then performs attention over the profile to find relevant lines from the profile to combine with the input, and then finally predicts the next utterance. We use the same representation and loss as in the Starspace model, so without the profile, the two models are identical.
When the profile is available attention is performed by computing the similarity of the input $q$ with the profile sentences $p_i$, computing the softmax, and taking the weighted sum:
\[
   q^+ = q + \sum s_i p_i,  ~~~~~~~~ s_i = {\mbox{Softmax}}(sim(q, p_i))
\]
where ${\mbox{Softmax}}(z_i) = e^{z_i}/\sum_j e^{z_j}$.
One can then rank the candidates $c'$ using $sim(q^+,c')$.
One can also perform multiple ``hops'' of attention over the profile rather than one, as shown here, although that did not bring significant gains in our parameter sweeps. 

\subsection{Key-Value Profile Memory Network}\label{sec:kvmem}

The key-value (KV) memory network \cite{miller2016key} was proposed as an improvement to the memory network by performing attention over keys and outputting the values (instead of the same keys as in the original), which can outperform memory networks dependent on the task and definition of the key-value pairs. Here, we apply this model to dialogue, and consider the keys as dialog histories (from the training set), and the values as the next dialogue utterances, i.e., the replies from the speaking partner. 
This allows the model to have a memory of past dialogues that it can directly use to help influence its prediction for the current conversation. 
The model we choose is identical to the profile memory network just described in the first hop over profiles, while in the second hop, $q^+$ is used to attend over the keys and output a weighted sum of values as before, producing 
$q^{++}$. This is then used to rank the candidates $c'$  using $sim(q^{++},c')$ as before.  As the set of (key-value) pairs is large this would make training very slow. In our experiments we simply trained the profile memory network and used the same weights from that model and applied this architecture at test time instead. Training the model directly would presumably give better results, however this heuristic already proved beneficial compared to the original network. 

%

\subsection{Seq2Seq}

The input sequence $x$ is encoded by applying $h^e_t = LSTM_{enc}(x_t \mid h^e_{t-1})$. We use GloVe \citep{pennington2014glove} for our word embeddings. The final hidden state, $h^e_t$, is fed into the decoder $LSTM_{dec}$ as the initial state $h^d_0$. For each time step $t$, the decoder then produces the probability of a word $j$ occurring in that place via the softmax, i.e.,
\begin{equation*}
p(y_{t,j} = 1 \mid y_{t-1}, \ldots, y_1) = \frac{\exp (w_j h^d_t)}{\sum^K_{j'=1} \exp (w_{j'} h^d_t)}.
\end{equation*}
The model is trained via negative log likelihood. The basic model can be extended to include persona information, in which case we simply prepend it to the input sequence $x$, i.e., $x =  \forall p \in P \mid\mid x$, where $\mid\mid$ denotes concatenation.
For the OpenSubtitles and Twitter datasets trained in Section \ref{sec:human-eval} we found training a language model (LM), essentially just the decoder part of this model, worked better and we report that instead.

\subsection{Generative Profile Memory Network}

Finally, we introduce a generative model that encodes each of the profile entries as individual memory representations in a memory network. As before, the dialogue history is encoded via $LSTM_{enc}$, the final state of which is used as the initial hidden state of the decoder. Each entry $p_i = \langle p_{i,1}, \ldots, p_{i,n} \rangle \in P$ is then encoded via $f(p_i) = \sum_j^{|p_i|} \alpha_i p_{i,j}$. That is, we weight words by their inverse term frequency: $\alpha_i = 1 / (1 + log(1 + \text{tf}))$ where $\text{tf}$ is computed from the GloVe index via Zipf's law\footnote{$\text{tf} = 1e6 * 1/(idx^{1.07})$}. Let $F$ be the set of encoded memories. The decoder now attends over the encoded profile entries, i.e., we compute the mask $a_t$, context $c_t$ and next input $\hat{x}_t$ as:
\begin{eqnarray*}
a_t = softmax(F W_a h^d_t), \\
c_t = a_t^\intercal F; ~~\hat{x}_t = tanh(W_c [c_{t-1}, x_t]).
\end{eqnarray*}
\ifarxiv
This model is illustrated in Figure \ref{fig:PMN-gen}.
\fi
If the model has no profile information, and hence no memory, it becomes equivalent to the Seq2Seq model.

\begin{table*}[t]
  \centering
  \begin{tabular}{lcccccc}
  \toprule
  \multirow{2}{*}{\textbf{Method}} & \multicolumn{2}{c}{\textbf{No Persona}} & \multicolumn{2}{c}{\textbf{Original Persona}} & \multicolumn{2}{c}{\textbf{Revised Persona}}\\
   & \textbf{ppl} & \textbf{hits@1} & \textbf{ppl} & \textbf{hits@1}  &\textbf{ppl} & \textbf{hits@1} 
   \\
  \midrule
  {\em Generative Models} \\
  Seq2Seq        & 38.08 & 0.092 & 40.53 & 0.084 &  40.65 & 0.082 \\
  Profile Memory & 38.08 & 0.092 & 34.54 & 0.125 & 38.21 & 0.108 \\
  \midrule
  {\em Ranking Models} \\
    IR baseline        & - & 0.214 &  - & 0.410  & - &  0.207  \\
  Starspace          & - & 0.318 &   - & 0.491  & - & 0.322   \\
 Profile Memory    & - & 0.318  &  - & 0.509  &  - & 0.354  \\
 KV Profile Memory & - & 0.349  &  - & 0.511  &  - & 0.351   \\
  \bottomrule
  \end{tabular}
  \caption{{\bf Evaluation of dialog utterance prediction with various models} in three settings: without conditioning on a persona, conditioned on the speakers given persona (``Original Persona''),  or a revised persona that does not have word overlap.      \label{tab:all-results}
     }
\end{table*}

\begin{table*}[t]
  \centering
  \begin{tabular}{ll|cccc}
  \toprule
  \multicolumn{2}{c|}{{\bf Method}} & & & &  \textbf{Persona}   \\
   Model & Profile & \textbf{Fluency} &  \textbf{Engagingness} & \textbf{Consistency} & 
   \textbf{Detection}   \\
  \midrule
  Human  & Self  & 4.31(1.07) & 4.25(1.06)   & 4.36(0.92) &  0.95(0.22) \\
  \midrule
  {\em Generative PersonaChat Models} & \\
  Seq2Seq & None              & 3.17(1.10)&  3.18(1.41) & 2.98(1.45) & 0.51(0.50)\\ %
  Profile Memory   & Self   &3.08(1.40) &  3.13(1.39) & 3.14(1.26) & 0.72(0.45) \\
  \midrule
  {\em Ranking PersonaChat Models} & \\
  KV Memory   & None      & 3.81(1.14) & 3.88(0.98) & 3.36(1.37) & 0.59(0.49)\\ 
  KV Profile Memory   & Self & 3.97(0.94) & 3.50(1.17)  & 3.44(1.30) & 0.81(0.39)\\ 
  \midrule
   Twitter LM & None & 3.21(1.54) & 1.75(1.04) & 1.95(1.22) & 0.57(0.50) \\
   OpenSubtitles 2018 LM & None & 2.85(1.46) & 2.13(1.07) & 2.15(1.08) & 0.35(0.48) \\
   OpenSubtitles 2009 LM & None & 2.25(1.37) & 2.12(1.33) & 1.96(1.22) & 0.38(0.49) \\
   OpenSubtitles 2009 KV Memory & None & 2.14(1.20)  & 2.22(1.22) & 2.06(1.29)  & 0.42(0.49) \\ 
  \bottomrule
  \end{tabular}
 \caption{{\bf Human Evaluation} of  various {\sc persona-chat} models,    along with a  comparison to human performance, and Twitter and OpenSubtitles based models (last 4 rows), standard deviation in parenthesis.
     \label{tab:human-eval}
     } 
\end{table*}

\section{Experiments}

We first report results using automated evaluation metrics, and subsequently perform an extrinsic evaluation where crowdsourced workers perform a human evaluation of our models. 

\subsection{Automated metrics}

The main results are reported in Table \ref{tab:all-results}.
%
Overall, the results show the following key points:

{\bf Persona Conditioning} Most models improve significantly when conditioning prediction on their own persona 
 at least for the original (non-revised) versions, which is an easier task than the revised ones which have no word overlap. For example, the Profile Memory generation model has improved perplexity and hits@1 compared to Seq2Seq, and all the ranking algorithms (IR baseline, Starspace and Profile Memory Networks) obtain improved hits@1.

{\bf Ranking vs. Generative.} Ranking models are far better than generative models at ranking. This is perhaps obvious as that is the metric they are optimizing, but still the performance difference is quite stark. 
It may be that the word-based probability which generative models use works well, but is not calibrated well enough to give a sentence-based probability which ranking requires.
Human evaluation is also used to compare these methods, which we perform in Sec. \ref{sec:human-eval}.

{\bf Ranking Models.} For the ranking models, the IR baseline is outperformed by Starspace due to its learnt similarity metric, which in turn is outperformed by Profile Memory networks due to the attention mechanism over the profiles (as all other parts of the models are the same). Finally KV Profile Memory networks outperform Profile Memory Networks in the no persona case due to the ability to consider neighboring dialogue history and next utterance pairs in the training set that are similar to the current dialogue, however when using persona information the performance is similar.

{\bf Revised Personas.} Revised personas are much harder to use. We do however still see some gain for the Profile Memory networks compared to none (0.354 vs. 0.318 hits@1). 
We also tried two variants of training: with the original personas in the training set or the revised ones, a comparison of which is shown in Table \ref{tab:retrieval-results} of the Appendix.  
{\em Training} on revised personas helps, both for test examples that are in original form or revised form, likely due to the model be forced to learn more than simple word overlap, forcing the model to generalize more (i.e., learn semantic similarity of differing phrases).

{\bf Their Persona.} We can also condition a model on the other speaker's persona, or both personas
at once, the results of which are in Tables \ref{tab:generative-results}
and \ref{tab:retrieval-results} in the Appendix.
Using ``Their persona'' has less impact on this dataset. We believe this is because most speakers tend to focus on themselves when it comes to their interests. It would be interesting how often this is the case in other datasets. Certainly this is skewed by the particular instructions one could give to the crowdworkers. For example if we gave the instructions ``try not to talk about yourself, but about the other's interests' likely these metrics would change.


\subsection{Human Evaluation} \label{sec:human-eval}

As automated metrics are notoriously poor for evaluating dialogue \citep{liu2016not} we also perform human evaluation using crowdsourced workers.
The procedure is as follows. We perform almost exactly the same setup as in the dataset collection process itself as in Section \ref{personachatter}. In that setup,  we paired two Turkers and assigned them each a random (original) persona from the collected pool, and asked them to chat. Here, from the Turker's point of view everything looks the same except instead of being paired with a Turker they are paired with one of our models instead (they do not know this). In this setting, for both the Turker and the model, the personas come from the test set pool.

After the dialogue, we then ask the Turker some additional questions in order to evaluate the quality of the model. 
\ifarxiv
They are, in order:
\begin{itemize}

\item {\bf Fluency}: We ask them to judge the fluency of the other speaker as a score from 1 to 5, where 1 is ``not fluent at all'', 5 is ``extremely fluent'', and 3 is ``OK''. 

\item {\bf Engagingness}: We ask them to judge the engagingness of the other speaker {\em disregarding fluency} from 1-5, where 1 is ``not engaging at all'', 5 is ``extremely engaging'', and 3 is ``OK''.

\item {\bf Consistency}: We ask them to judge the consistency of the persona of the other speaker, where we give the example that ``I have a dog''  followed by ``I have no pets'' is not consistent. The score is again from 1-5.

\item {\bf Profile Detection}: Finally, we display two possible profiles, and ask which is more likely to be the profile of the person the Turker just spoke to. One profile is chosen at random, and the other is the true persona given to the model.
\end{itemize}
\else
We ask them to evaluate fluency, engagingness and consistency (scored between 1-5). Finally, we measure the ability to detect the other speaker's profile by displaying two possible profiles, and ask which is more likely to be the profile of the person the Turker just spoke to.  More details of these measures are given in the Appendix.
\fi

The results are reported in Table \ref{tab:human-eval} for the best performing generative and ranking models, in both the No Persona and Self Persona categories, 100 dialogues each. We also evaluate the scores of human performance by replacing the chatbot with a human (another Turker). This effectively gives us upper bound scores which we can aim for with our models. Finally, and importantly, we compare our models trained on {\sc persona-chat} with chit-chat models trained with the Twitter and OpenSubtitles datasets (2009 and 2018 versions) instead, following \newcite{vinyals2015neural}. Example chats from a few of the models are shown in the Appendix in
Tables \ref{table:os-example}, \ref{table:s2s-example}, \ref{table:kvp-example}, \ref{table:gpm-example},
 \ref{table:opensubtitles2018-example} and \ref{table:twitter-example}.

Firstly, we see a difference in fluency, engagingness and consistency between all {\sc persona-chat}  models and the models trained on OpenSubtitles and Twitter. 
{\sc persona-chat} is a resource that is particularly strong at providing training data for the beginning of conversations, when the two speakers do not know each other, focusing on asking and answering questions, in contrast to other resources.
We also see suggestions of more subtle differences between the models, although these differences are obscured by the high variance of the human raters' evaluations. 
For example, in both the generative and ranking model cases, models endowed with a persona can be detected by the human conversation partner, as evidenced by the persona detection accuracies, whilst maintaining fluency and consistency compared to their non-persona driven counterparts.

 
Finding the balance between fluency, engagement, consistency, and a persistent persona remains a strong challenge for future research.

\subsection{Profile Prediction}

Two tasks could naturally be considered using {\sc PersonaChat}:
(1) next utterance prediction during dialogue, and (2) profile prediction given dialogue history. 
The main study of this work has been Task 1, where we have shown the use of profile information.
Task 2, however, can be used to extract such information.
While a full study is beyond the scope of this paper, we conducted some preliminary experiments,
the details of which are in Appendix \ref{app:profile-pred}.
They show (i) human speaker's profiles can be predicted from their dialogue with high accuracy
(94.3\%, similar to human performance in Table \ref{tab:human-eval})
 or even from the model's dialogue (23\% with KV Profile Memory) 
showing the model is paying attention to the 
human's interests. Further, the accuracies clearly improve with further dialogue, as shown in Table 
\ref{tab:task2b}. Combining Task 1 and Task 2 into a full system is an exciting area of 
future research.


\section{Conclusion \& Discussion}

In this work we have introduced the {\sc persona-chat} dataset, which consists of crowd-sourced dialogues where each participant plays the part of an assigned persona; and each (crowd-sourced) persona has a word-distinct paraphrase.  We test various baseline models on this dataset, and show that models that have access to their own personas in addition to the state of the dialogue are scored as more consistent by annotators, although not more engaging.   On the other hand, we show that models trained on {\sc persona-chat} (with or without personas) are more engaging than models trained on dialogue from other resources (movies, Twitter).

We believe {\sc persona-chat} will be a useful resource for training components of future dialogue systems.  Because we have paired human generated profiles and conversations, the data aids the construction of agents that have consistent personalities and viewpoints.  Furthermore, predicting the profiles from a conversation moves chit-chat tasks in the direction of goal-directed dialogue, which has metrics for success. Because we collect paraphrases of the profiles, they cannot be trivially matched; indeed, we believe the original and rephrased profiles are interesting as a semantic similarity dataset in their own right.  We hope that the data will aid training agents that can ask questions about users' profiles, remember the answers, and use them naturally in conversation.  




\bibliography{iclr2018_conference}

\begin{thebibliography}{28}
\expandafter\ifx\csname natexlab\endcsname\relax\def\natexlab#1{#1}\fi

\bibitem[{Bordes et~al.(2016)}]{bordes2016learning}
Antoine Bordes, Y-Lan Boureau and Jason Weston. 2016.
\newblock Learning end-to-end goal-oriented dialog.
\newblock \emph{arXiv preprint arXiv:1605.07683}.

\bibitem[{Chen et~al.(2017)Chen, Fisch, Weston, and Bordes}]{chen2017reading}
Danqi Chen, Adam Fisch, Jason Weston, and Antoine Bordes. 2017.
\newblock Reading wikipedia to answer open-domain questions.
\newblock \emph{arXiv preprint arXiv:1704.00051}.

\bibitem[{Dodge et~al.(2015)Dodge, Gane, Zhang, Bordes, Chopra, Miller, Szlam,
  and Weston}]{dodge2015evaluating}
Jesse Dodge, Andreea Gane, Xiang Zhang, Antoine Bordes, Sumit Chopra, Alexander
  Miller, Arthur Szlam, and Jason Weston. 2015.
\newblock Evaluating prerequisite qualities for learning end-to-end dialog
  systems.
\newblock \emph{arXiv preprint arXiv:1511.06931}.

\bibitem[{Dunbar et~al.(1997)Dunbar, Marriott, and Duncan}]{dunbar1997human}
Robin~IM Dunbar, Anna Marriott, and Neil~DC Duncan. 1997.
\newblock Human conversational behavior.
\newblock \emph{Human nature}, 8(3):231--246.

\bibitem[{Joshi et~al.(2017)Joshi, Mi, and Faltings}]{joshi2017personalization}
Chaitanya~K Joshi, Fei Mi, and Boi Faltings. 2017.
\newblock Personalization in goal-oriented dialog.
\newblock \emph{arXiv preprint arXiv:1706.07503}.

\bibitem[{Li et~al.(2015)Li, Galley, Brockett, Gao, and
  Dolan}]{li2015diversity}
Jiwei Li, Michel Galley, Chris Brockett, Jianfeng Gao, and Bill Dolan. 2015.
\newblock A diversity-promoting objective function for neural conversation
  models.
\newblock \emph{arXiv preprint arXiv:1510.03055}.

\bibitem[{Li et~al.(2016{\natexlab{a}})Li, Galley, Brockett, Spithourakis, Gao,
  and Dolan}]{li2016persona}
Jiwei Li, Michel Galley, Chris Brockett, Georgios~P Spithourakis, Jianfeng Gao,
  and Bill Dolan. 2016{\natexlab{a}}.
\newblock A persona-based neural conversation model.
\newblock \emph{arXiv preprint arXiv:1603.06155}.

\bibitem[{Li et~al.(2016{\natexlab{b}})Li, Monroe, Ritter, Galley, Gao, and
  Jurafsky}]{li2016deep}
Jiwei Li, Will Monroe, Alan Ritter, Michel Galley, Jianfeng Gao, and Dan
  Jurafsky. 2016{\natexlab{b}}.
\newblock Deep reinforcement learning for dialogue generation.
\newblock \emph{arXiv preprint arXiv:1606.01541}.

\bibitem[{Liu et~al.(2016)Liu, Lowe, Serban, Noseworthy, Charlin, and
  Pineau}]{liu2016not}
Chia-Wei Liu, Ryan Lowe, Iulian~V Serban, Michael Noseworthy, Laurent Charlin,
  and Joelle Pineau. 2016.
\newblock How not to evaluate your dialogue system: An empirical study of
  unsupervised evaluation metrics for dialogue response generation.
\newblock \emph{arXiv preprint arXiv:1603.08023}.

\bibitem[{Lowe et~al.(2015)Lowe, Pow, Serban, and Pineau}]{lowe2015ubuntu}
Ryan Lowe, Nissan Pow, Iulian Serban, and Joelle Pineau. 2015.
\newblock The ubuntu dialogue corpus: A large dataset for research in
  unstructured multi-turn dialogue systems.
\newblock \emph{arXiv preprint arXiv:1506.08909}.

\bibitem[{Lucas et~al.(2009)Lucas, Fern{\'a}ndez, Salazar, Ferreiros, and
  San~Segundo}]{lucas2009managing}
JM~Lucas, F~Fern{\'a}ndez, J~Salazar, J~Ferreiros, and R~San~Segundo. 2009.
\newblock Managing speaker identity and user profiles in a spoken dialogue
  system.
\newblock \emph{Procesamiento del Lenguaje Natural}, 43:77--84.

\bibitem[{Miller et~al.(2016)Miller, Fisch, Dodge, Karimi, Bordes, and
  Weston}]{miller2016key}
Alexander Miller, Adam Fisch, Jesse Dodge, Amir-Hossein Karimi, Antoine Bordes,
  and Jason Weston. 2016.
\newblock Key-value memory networks for directly reading documents.
\newblock \emph{arXiv preprint arXiv:1606.03126}.

\bibitem[{Naaman et~al.(2010)Naaman, Boase, and Lai}]{naaman2010really}
Mor Naaman, Jeffrey Boase, and Chih-Hui Lai. 2010.
\newblock Is it really about me?: message content in social awareness streams.
\newblock In \emph{Proceedings of the 2010 ACM conference on Computer supported
  cooperative work}, pages 189--192. ACM.

\bibitem[{Pennington et~al.(2014)Pennington, Socher, and
  Manning}]{pennington2014glove}
Jeffrey Pennington, Richard Socher, and Christopher Manning. 2014.
\newblock Glove: Global vectors for word representation.
\newblock In \emph{Proceedings of the 2014 conference on empirical methods in
  natural language processing (EMNLP)}, pages 1532--1543.

\bibitem[{Rajpurkar et~al.(2016)Rajpurkar, Zhang, Lopyrev, and
  Liang}]{rajpurkar2016squad}
Pranav Rajpurkar, Jian Zhang, Konstantin Lopyrev, and Percy Liang. 2016.
\newblock Squad: 100,000+ questions for machine comprehension of text.
\newblock \emph{arXiv preprint arXiv:1606.05250}.

\bibitem[{Serban et~al.(2017{\natexlab{a}})Serban, Sankar, Germain, Zhang, Lin,
  Subramanian, Kim, Pieper, Chandar, Ke et~al.}]{serban2017deep}
Iulian~V Serban, Chinnadhurai Sankar, Mathieu Germain, Saizheng Zhang, Zhouhan
  Lin, Sandeep Subramanian, Taesup Kim, Michael Pieper, Sarath Chandar,
  Nan~Rosemary Ke, et~al. 2017{\natexlab{a}}.
\newblock A deep reinforcement learning chatbot.
\newblock \emph{arXiv preprint arXiv:1709.02349}.

\bibitem[{Serban et~al.(2015)Serban, Lowe, Charlin, and
  Pineau}]{serban2015survey}
Iulian~Vlad Serban, Ryan Lowe, Laurent Charlin, and Joelle Pineau. 2015.
\newblock A survey of available corpora for building data-driven dialogue
  systems.
\newblock \emph{arXiv preprint arXiv:1512.05742}.

\bibitem[{Serban et~al.(2016)Serban, Lowe, Charlin, and
  Pineau}]{serban2016generative}
Iulian~Vlad Serban, Ryan Lowe, Laurent Charlin, and Joelle Pineau. 2016.
\newblock Generative deep neural networks for dialogue: A short review.
\newblock \emph{arXiv preprint arXiv:1611.06216}.

\bibitem[{Serban et~al.(2017{\natexlab{b}})Serban, Sordoni, Lowe, Charlin,
  Pineau, Courville, and Bengio}]{serban2017hierarchical}
Iulian~Vlad Serban, Alessandro Sordoni, Ryan Lowe, Laurent Charlin, Joelle
  Pineau, Aaron~C Courville, and Yoshua Bengio. 2017{\natexlab{b}}.
\newblock A hierarchical latent variable encoder-decoder model for generating
  dialogues.

\bibitem[{Sordoni et~al.(2015)Sordoni, Galley, Auli, Brockett, Ji, Mitchell,
  Nie, Gao, and Dolan}]{sordoni2015neural}
Alessandro Sordoni, Michel Galley, Michael Auli, Chris Brockett, Yangfeng Ji,
  Margaret Mitchell, Jian-Yun Nie, Jianfeng Gao, and Bill Dolan. 2015.
\newblock A neural network approach to context-sensitive generation of
  conversational responses.
\newblock \emph{arXiv preprint arXiv:1506.06714}.

\bibitem[{Sukhbaatar et~al.(2015)Sukhbaatar, Weston, Fergus et~al.}]{memn2n}
Sainbayar Sukhbaatar, Jason Weston, Rob Fergus, et~al. 2015.
\newblock End-to-end memory networks.
\newblock In \emph{Advances in neural information processing systems}, pages
  2440--2448.

\bibitem[{Sutskever et~al.(2014)Sutskever, Vinyals, and
  Le}]{sutskever2014sequence}
Ilya Sutskever, Oriol Vinyals, and Quoc~V Le. 2014.
\newblock Sequence to sequence learning with neural networks.
\newblock In \emph{Advances in neural information processing systems}, pages
  3104--3112.

\bibitem[{Vinyals and Le(2015)}]{vinyals2015neural}
Oriol Vinyals and Quoc Le. 2015.
\newblock A neural conversational model.
\newblock \emph{arXiv preprint arXiv:1506.05869}.

\bibitem[{Voorhees et~al.(1999)}]{voorhees1999trec}
Ellen~M Voorhees et~al. 1999.
\newblock The trec-8 question answering track report.
\newblock In \emph{Trec}, volume~99, pages 77--82.

\bibitem[{Weizenbaum(1966)}]{weizenbaum1966eliza}
Joseph Weizenbaum. 1966.
\newblock Eliza—a computer program for the study of natural language
  communication between man and machine.
\newblock \emph{Communications of the ACM}, 9(1):36--45.

\bibitem[{Wu et~al.(2017)Wu, Fisch, Chopra, Adams, Bordes, and
  Weston}]{wu2017starspace}
Ledell Wu, Adam Fisch, Sumit Chopra, Keith Adams, Antoine Bordes, and Jason
  Weston. 2017.
\newblock Starspace: Embed all the things!
\newblock \emph{arXiv preprint arXiv:1709.03856}.

\bibitem[{Young et~al.(2013)Young, Ga{\v{s}}i{\'c}, Thomson, and
  Williams}]{young2013pomdp}
Steve Young, Milica Ga{\v{s}}i{\'c}, Blaise Thomson, and Jason~D Williams.
  2013.
\newblock Pomdp-based statistical spoken dialog systems: A review.
\newblock \emph{Proceedings of the IEEE}, 101(5):1160--1179.

\bibitem[{Young(2000)}]{young2000probabilistic}
Steve~J Young. 2000.
\newblock Probabilistic methods in spoken--dialogue systems.
\newblock \emph{Philosophical Transactions of the Royal Society of London A:
  Mathematical, Physical and Engineering Sciences}, 358(1769):1389--1402.

\end{thebibliography}
\bibliographystyle{acl_natbib}

\newpage
\clearpage
\newpage
\clearpage
\appendix


\begin{table*}[t]
  \centering
  \begin{tabular}{llllllll}
  \toprule
  \multirow{2}{*}{\textbf{Persona}} & \multirow{2}{*}{\textbf{Method}} & \multicolumn{3}{c}{\textbf{Original}} & \multicolumn{3}{c}{\textbf{Revised}}\\
  & & \textbf{ppl} & \textbf{hits@1} & \textbf{F1}&\textbf{ppl} & \textbf{hits@1} & \textbf{F1}\\
  \midrule
  No Persona & & 38.08 & 0.092 & 0.168&38.08 & 0.092&0.168\\\midrule
  \multirow{3}{*}{Self Persona} & Seq2Seq & 40.53 & 0.084 &\textbf{0.172}& 40.65  & 0.082&\textbf{0.171}\\
   & Profile Memory & \textbf{34.54} & \textbf{0.125} &\textbf{0.172}& 38.21 & \textbf{0.108}&0.170\\\midrule
  \multirow{3}{*}{Their Persona} & Seq2Seq & 41.48 & 0.075 &0.168& 41.95 & 0.074&0.168\\
   & Profile Memory & 36.42 & 0.105 &0.167& \textbf{37.75} & 0.103&0.167\\\midrule
  \multirow{3}{*}{Both Personas} & Seq2Seq & 40.14 & 0.084 &0.169& 40.53 & 0.082&0.166\\
   & Profile Memory & 35.27 & 0.115 &0.171& 38.48 & 0.106&0.168\\ 
  \bottomrule
  \end{tabular}
  \caption{{\bf Evaluation of dialog utterance prediction with generative models} in four settings: conditioned on the speakers persona (``self persona''), the dialogue partner's persona (``their persona''), both or none. The personas are either the original source given to Turkers to condition the dialogue, or the revised personas that do not have word overlap. In the ``no persona'' setting, the models are equivalent, so we only report once.
     \label{tab:generative-results}
     }
\end{table*}

\begin{table*}[t]
  \begin{center}
      \begin{tabular}{l|cc|cc|cc|cc }
      \toprule
      ~&  \multicolumn{2}{c}{No Persona} & \multicolumn{2}{|c}{Self Persona} & \multicolumn{2}{|c}{Their Persona} & \multicolumn{2}{|c}{Both Personas} \\ 
      Method & Orig & Rewrite & Orig & Rewrite & Orig & Rewrite & Orig & Rewrite \\ 
      \midrule
      IR baseline  &0.214 & 0.214 & 0.410 & 0.207  &  0.181  & 0.181 & 0.382& 0.188 \\
      \multicolumn{8}{l}{{\em Training on original personas}}\\
      Starspace    & 0.318 & 0.318 &  0.481  & 0.295& 0.245 & 0.235 & 0.429 & 0.258\\
      Profile Memory        &  0.318 & 0.318 & 0.473 & 0.302 & 0.283 & 0.267 & 0.438 & 0.266\\    
      \multicolumn{8}{l}{{\em Training on revised personas}}\\
      Starspace    &  0.318 & 0.318 & 0.491 & 0.322 & 0.271 & 0.261 & 0.432 & 0.288\\
      Profile Memory        &  0.318 & 0.318 & 0.509 & 0.354 & 0.299 & 0.294 & 0.467 & 0.331\\
      KV Profile Memory     &  0.349 & 0.349 & 0.511 & 0.351 & 0.291 &  0.289      &          0.467 &  0.330 \\
      \bottomrule
      \end{tabular}
      \caption{{\bf Evaluation of dialog utterance prediction with ranking models} using hits@1 in four settings: conditioned on the speakers persona ("self persona"), the dialogue partner's persona ("their persona"), both or none. The personas are either the original source given to Turkers to condition the dialogue, or the rewritten personas that do not have word overlap, explaining the poor performance of IR in that case. }
      \label{tab:retrieval-results}
  \end{center}
\end{table*}

\section{Next Utterance Prediction Additional Evaluation Metrics}

In Table \ref{tab:generative-results} and Table \ref{tab:retrieval-results} we show additional results on next utterance prediction.
In particular we give results for conditioning on the other person's persona ("Their") instead of one's own ("Self") or the combination ("Both"). In particular, we see that, knowing the other's persona does not help precision on this data with these models. Finally, we show in  Table \ref{tab:retrieval-results} for the ranking models the performance difference from training on original personas versus revised ones. Revised personas give improved results, perhaps because the models are forced to learn more than trivial word overlap (i.e. semantic similarity of differing phrases).

\section{Example Dialogs between Humans and Models}

In Tables \ref{table:os-example}, \ref{table:s2s-example}, \ref{table:kvp-example}, 
 \ref{table:opensubtitles2018-example}, \ref{table:twitter-example}
we show  example dialogues between Turkers
and various models that were collected as part of the human evaluation of Section \ref{sec:human-eval}.

\begin{table*}[t]
  \begin{center}
    \begin{small}
      \begin{tabular}{l|l}
        \toprule
        \textbf{Persona 1} & \textbf{Persona 2 (Not used by model)}\\
        \midrule
I am a doctor &  I eat sardines for breakfast daily \\
I have one brother & I just had surgery\\
My mom is a secretary & I love playing Frisbee \\
I like to eat a lot of meat & I have two large dogs \\
I am a bodybuilder & I never finished high school \\
\bottomrule
\multicolumn{2}{l}{ }\\
\multicolumn{2}{l}{[PERSON 1:] Hi I am Dr. Jenson. What is your name?}\\
\multicolumn{2}{l}{[PERSON 2:] Miss Dawes.}\\
\multicolumn{2}{l}{[PERSON 1:] Hi Miss Dawes, tell me about your family!}\\
\multicolumn{2}{l}{[PERSON 2:] Miss Baggoli, I know it must show on your records that my name is Mary, but everyone} \\\multicolumn{2}{l}{has referred to me as Lola since I was a squealing infant in my mother' s arms. }\\
\multicolumn{2}{l}{[PERSON 1:] fascinating. My secretary mother would have made note of all of that.}\\
\multicolumn{2}{l}{[PERSON 2:] In addition to the remainder, it writes my autobiography.}\\
\multicolumn{2}{l}{[PERSON 1:] Oh yeah? well I enjoy a nice steak. how about you?}\\
\multicolumn{2}{l}{[PERSON 2:] Father always used to say that if we ever had the money you have...}\\ \multicolumn{2}{l}{we' d have steak and ice cream three times every day!}\\
\multicolumn{2}{l}{[PERSON 1:] I eat all the steak and chicken too, even bacon}\\
\multicolumn{2}{l}{[PERSON 2:] Father always used to say that if we ever had the money you have...}\\ \multicolumn{2}{l}{we' d have steak and ice cream three times every day!}\\
\multicolumn{2}{l}{[PERSON 1:] yeah you said that already}\\
\multicolumn{2}{l}{[PERSON 2:] Dynamic structure can absorb shock.}\\
\multicolumn{2}{l}{[PERSON 1:] yes it can. I know about that as a bodybuilder}\\
\multicolumn{2}{l}{[PERSON 2:] Ladies and gentlemen... the strongest man in the world is blowing up a hot water bottle.}\\
      \end{tabular}
      \caption{Example dialog between a human (Person 1) and the OpenSubtitles KV Memory Network model (Person 2). 
 \label{table:os-example}}
    \end{small}
  \end{center}
\end{table*}

\begin{table*}[t]
  \begin{center}
    \begin{small}
      \begin{tabular}{l|l}
        \toprule
        \textbf{Persona 1} & \textbf{Persona 2 (Not used by model)}\\
        \midrule
I am afraid of dogs & I am a carpenter \\
I used to work at a carnival & I used to drink but I stopped five years ago\\
I like to drink scotch to relax & I have a wife and three kids\\
I write short stories for a living & I like playing poker\\
I like dark superhero movies & I dont have many friends\\
\bottomrule
\multicolumn{2}{l}{ }\\
\multicolumn{2}{l}{[PERSON 1:] Hi are you relaxing on this fine evening?}\\
\multicolumn{2}{l}{[PERSON 2:] I am doing good and you?}\\
\multicolumn{2}{l}{[PERSON 1:] I am well, just drinking some scotch and working on my writing}\\
\multicolumn{2}{l}{[PERSON 2:] I love to cook and bake but I rarely make time for it} \\
\multicolumn{2}{l}{[PERSON 1:] what do you usually bake?}\\
\multicolumn{2}{l}{[PERSON 2:] I do not cook, I bake cakes at a local restaurant. how about you?}\\
\multicolumn{2}{l}{[PERSON 1:] I used to be a juggler at a carnival but now I am an author}\\
\multicolumn{2}{l}{[PERSON 2:] I am a dancer, what do you do?}\\ 
\multicolumn{2}{l}{[PERSON 1:] I just told you. Do you have any pets?}\\
\multicolumn{2}{l}{[PERSON 2:] no, I am allergic to pets unfortunately.}\\ 
\multicolumn{2}{l}{[PERSON 1:] I understand. I am not allergic but I am terrified of dogs!}\\
\multicolumn{2}{l}{[PERSON 2:] I have a cat. she has been with me through three wives, haha.}\\
\multicolumn{2}{l}{[PERSON 1:] Funny that you just told me you have no pets}\\
\multicolumn{2}{l}{[PERSON 2:] I do not have any animals.}\\
      \end{tabular}
      \caption{Example dialog between a human (Person 1) and the Seq2Seq model (Person 2). 
 \label{table:s2s-example}}
    \end{small}
  \end{center}
\end{table*}

\begin{table*}[t]
  \begin{center}
    \begin{small}
      \begin{tabular}{l|l}
        \toprule
        \textbf{Persona 1} & \textbf{Persona 2}\\
        \midrule
I am engaged to be married & I am a writer \\ 
I used to play sports in high school & I write romance novels\\
My favorite sport is soccer &  I live on a faraway island \\
I am a college graduate& I am an introvert \\
I love watching sports on TV& I have two cats \\
\bottomrule
\multicolumn{2}{l}{ }\\
\multicolumn{2}{l}{[PERSON 1:] I just got done playing soccer. What have you been up to?}\\
\multicolumn{2}{l}{[PERSON 2:] I do not have pets. I spend my free time writing. what about you?}\\
\multicolumn{2}{l}{[PERSON 1:] Just graduated from college. Looking for work. }\\
\multicolumn{2}{l}{[PERSON 2:] yes I graduated, in writing and poetry}\\
\multicolumn{2}{l}{[PERSON 1:] Have you had anything published?}\\
\multicolumn{2}{l}{[PERSON 2:] I mainly write fiction novels.}\\
\multicolumn{2}{l}{[PERSON 1:] Do you like to watch sports?}\\
\multicolumn{2}{l}{[PERSON 2:] do you like kings of leon my favorite by them is use somebody}\\
\multicolumn{2}{l}{[PERSON 1:] Are you married? I will be married soon. }\\
\multicolumn{2}{l}{[PERSON 2:] haha, no time. I have got a novel to finish.}\\
\multicolumn{2}{l}{[PERSON 1:] What is your novel about?}\\
\multicolumn{2}{l}{[PERSON 2:] it is a love story. I really enjoy writing. do you write or read?}\\
\multicolumn{2}{l}{[PERSON 1:] Like to read. Spend most time watching football. }\\
\multicolumn{2}{l}{[PERSON 2:] yes I do I am and author for a living}\\
      \end{tabular}
      \caption{Example dialog between a human (Person 1) and the Key-Value Profile Memory Network with Self Persona. 
 \label{table:kvp-example}}
    \end{small}
  \end{center}
\end{table*}

\begin{table*}[t]
  \begin{center}
    \begin{small}
      \begin{tabular}{l|l}
        \toprule
        \textbf{Persona 1} & \textbf{Persona 2}\\
        \midrule
I dont like exercise or physical activity & My favorite type of food is Italian \\ 
My favorite food is ice cream & I have a big pet dog\\
I am a homebody &  I want to travel to Europe \\
 I have a husband who I love very much & I love reading books about travel \\
I love watching movies and TV& I like to walk long distances \\
\bottomrule
\multicolumn{2}{l}{ }\\
\multicolumn{2}{l}{[PERSON 1:] Hello, how are you doing today?}\\
\multicolumn{2}{l}{[PERSON 2:] I am good. I just got finished walking my dog.}\\
\multicolumn{2}{l}{[PERSON 1:] Oh cool! I am just sitting watching a movie. }\\
\multicolumn{2}{l}{[PERSON 2:] I have been watching that one before. I love it.}\\
\multicolumn{2}{l}{[PERSON 1:] Me too. I don't like going out much so I enjoy movies and TV.}\\
\multicolumn{2}{l}{[PERSON 2:] that is a great idea. I am going to read a book on writing.}\\
\multicolumn{2}{l}{[PERSON 1:] Oh, are you a writer?}\\
\multicolumn{2}{l}{[PERSON 2:] yes, I love to read. I am a great author.}\\
\multicolumn{2}{l}{[PERSON 1:] What sorts of topics do you write about?}\\
\multicolumn{2}{l}{[PERSON 2:] I could read all of those. I am sure you enjoy reading.}\\
\multicolumn{2}{l}{[PERSON 1:] I do, but I prefer movies.}\\
\multicolumn{2}{l}{[PERSON 2:] I have not, I can not do that. I have to go to the east coast.}\\
\multicolumn{2}{l}{[PERSON 1:] Why can't you watch movies?}\\
\multicolumn{2}{l}{[PERSON 2:] I have not. I have always had that done.}\\
      \end{tabular}
      \caption{Example dialog between a human (Person 1) and the Generative Profile Memory Network with Self Persona. 
 \label{table:gpm-example}}
    \end{small}
  \end{center}
\end{table*}

\begin{table*}[t]
  \begin{center}
    \begin{small}
      \begin{tabular}{l|l}
        \toprule
        \textbf{Persona 1} & \textbf{Persona 2 (Not used by model)}\\
     \midrule   
I am very active & I am an engineering student \\ 
I live in the desert & I work at Eureka \\
I enjoy the outdoors & I like to celebrate with friends  \\
I was named something pretty common & I enjoy spending time with my toes in the sand\\
I instruct people how to get fit as a job &  I study to be a businessman \\
\bottomrule
\multicolumn{2}{l}{ }\\
\multicolumn{2}{l}{[PERSON 1:] Hello there, how are you?}\\
\multicolumn{2}{l}{[PERSON 2:] I'm fine.}\\
\multicolumn{2}{l}{[PERSON 1:] How is the weather, it is hot here }\\
\multicolumn{2}{l}{[PERSON 2:] It's so hot.}\\
\multicolumn{2}{l}{[PERSON 1:] oh yes tell me about it especially when you live in a desert.}\\
\multicolumn{2}{l}{[PERSON 2:] the sea is shining.}\\
\multicolumn{2}{l}{[PERSON 1:] you at the beach, i love the beach}\\
\multicolumn{2}{l}{[PERSON 2:] wait, wait, wait...}\\
\multicolumn{2}{l}{[PERSON 1:] what's up, you going somewhere }\\
\multicolumn{2}{l}{[PERSON 2:] you can be a good friend.}\\
\multicolumn{2}{l}{[PERSON 1:] yes, if you want me to}\\
\multicolumn{2}{l}{[PERSON 2:] I'm sorry.}\\
      \end{tabular}
      \caption{Example dialog between a human (Person 1) and the Language Model trained on the OpenSubtitles 2018 dataset (does not use persona). 
 \label{table:opensubtitles2018-example}}
    \end{small}
  \end{center}
\end{table*}

\begin{table*}[t]
  \begin{center}
    \begin{small}
      \begin{tabular}{l|l}
        \toprule
        \textbf{Persona 1} & \textbf{Persona 2 (Not used by model)}\\
     \midrule   
I love ordering from Amazon & I really enjoy watching classic films over the new ones \\ 
I am a mother & My mother is the greatest chef in the world \\
I have a very active imagination & I stand out among my surroundings  \\
Winter holidays have always been my favorite & My son and daughter lived with me full time after I\\
I love the Harry Potter books &  \hspace{2em} separated from my husband\\
\bottomrule
\multicolumn{2}{l}{ }\\
\multicolumn{2}{l}{[PERSON 1:] Good afternoon! How are you?}\\
\multicolumn{2}{l}{[PERSON 2:] i'm not sure. I was just talking about the other day.}\\
\multicolumn{2}{l}{[PERSON 1:] Do you like to read? I like reading Harry Potter. }\\
\multicolumn{2}{l}{[PERSON 2:] I love you so much}\\
\multicolumn{2}{l}{[PERSON 1:] That's funny. You don't know me.}\\
\multicolumn{2}{l}{[PERSON 2:] I think you have to be a good person.}\\
\multicolumn{2}{l}{[PERSON 1:] Where do you live? Are you in the US?}\\
\multicolumn{2}{l}{[PERSON 2:] yes, I just found out that there were only 10 people who were in the city.}\\
\multicolumn{2}{l}{[PERSON 1:] Do you have any children? I am a mother to 1 cat. }\\
\multicolumn{2}{l}{[PERSON 2:] this is the first time in history, but not a few.}\\
\multicolumn{2}{l}{[PERSON 1:] Is it cold where you are?}\\
\multicolumn{2}{l}{[PERSON 2:] I don't even know what I'm talking about.}\\
      \end{tabular}
      \caption{Example dialog between a human (Person 1) and the Language Model trained on the Twitter dataset (does not use persona). 
 \label{table:twitter-example}}
    \end{small}
  \end{center}
\end{table*}

\section{Human Evaluation Measures}

After dialogues between humans and a model, we then ask the Turker some additional questions in order to evaluate the quality of the model. 
They are, in order:
\begin{itemize}
\item {\bf Fluency}: We ask them to judge the fluency of the other speaker as a score from 1 to 5, where 1 is ``not fluent at all'', 5 is ``extremely fluent'', and 3 is ``OK''. 

\item {\bf Engagingness}: We ask them to judge the engagingness of the other speaker {\em disregarding fluency} from 1-5, where 1 is ``not engaging at all'', 5 is ``extremely engaging'', and 3 is ``OK''.

\item {\bf Consistency}: We ask them to judge the consistency of the persona of the other speaker, where we give the example that ``I have a dog''  followed by ``I have no pets'' is not consistent. The score is again from 1-5.

\item {\bf Profile Detection}: Finally, we display two possible profiles, and ask which is more likely to be the profile of the person the Turker just spoke to. One profile is chosen at random, and the other is the true persona given to the model.
\end{itemize}

\section{Profile Prediction}\label{app:profile-pred}

While the main study of this work is the ability to improve next utterance classification
by conditioning on a persona, 
one could naturally consider two tasks:
(1) next utterance prediction during dialogue, and (2) profile prediction given dialogue history. 
In the main paper we show that Task 1 can be improved by using profile information.
Task 2, however, can be used to extract such information.

In this section we conduct a preliminary study of the ability to predict the persona
of a speaker given a set of dialogue utterances. 
We consider the dialogues between humans (PERSON 0)  and our best performing model, the retrieval-based Key-Value Profile Memory Network (PERSON 1) from Section \ref{sec:human-eval}. 
We tested the ability to predict the profile information of the two speakers from the dialogue
utterances of each speaker, considering all four combinations.
We employ the same 
IR baseline model used in the main paper to predict profiles: it ranks profile candidates, either at the entire profile level (considering all the sentences that make up the profile as a bag) or at the  sentence level (each  sentence individually). 
We consider 100 negative profile candidates for each positive profile, and compute the error rate of
predicting the true profile averaged over all dialogues and candidates.
The results are given in Table \ref{tab:task2a},  both for the model conditioned on profile information, and the same KV Memory model that is not.
The results indicate the following:
\begin{itemize}
\item It is possible to predict the humans profile from their dialogue utterances
(PERSON 0, Profile 0) with high accuracy at both the profile and sentence level, independent of the model they speaking to.
\item Similarly the model's profile can be predicted with high accuracy from its utterances (PERSON 1, Profile 1) when it is conditioned on the profile, otherwise this is chance level (w/o Profile).
\item It is possible to predict the model's profile from the human's dialogue, but with a lower accuracy (PERSON 0, Profile 1) as long as the model is conditioned on its own profile. This indicates the human responds to the model's utterances and pays attention to the model's interests. 
\item Similarly, the human's profile can be predicted from the model's dialogue, but with lower accuracy. Interestingly, the model without profile conditioning is better at this, perhaps because it does not concentrate on talking about itself, and pays more attention to responding to the human's interests. There appears to be a tradeoff that needs to be explored and understood here.
\end{itemize}

We also study the performance of profile prediction as the dialogue progresses, by computing error
rates for dialogue lengths 1 to 8 (the longest length we consider in this work). 
The results, given in Table \ref{tab:task2b}, show the error rate of predicting the persona 
decreases in all cases as dialogue length increases.

Overall, the results in this section 
show that it is plausible to predict profiles given dialogue utterances, which is
an important extraction task. Note that better results could likely be achieved with more sophisticated models.

\begin{table*}[t]
  \centering
  \begin{tabular}{ll|ll|ll}
  \toprule
  \multirow{3}{*}{\textbf{Speaker}} & 
  \multirow{3}{*}{\textbf{Profile}} & 
   \multicolumn{2}{c}{\textbf{Profile Level}} &   \multicolumn{2}{c}{\textbf{Sentence Level}} \\
&   &  KV Profile & KV w/o Profile    &  KV Profile & KV w/o  Profile \\
  \midrule
PERSON 0 & Profile 0  &  0.057  &  0.017  & 0.173 & 0.141 \\
PERSON 0 & Profile 1  &  0.234  &  0.491 & 0.431 & 0.518 \\
PERSON 1 & Profile 0  &  0.254  &  0.112  & 0.431 & 0.349 \\
PERSON 1 & Profile 1  &  0.011  &  0.512  & 0.246 & 0.530 \\
  \bottomrule
  \end{tabular}
  \caption{{\bf Profile Prediction.}
     \label{tab:task2a}
  Error rates are given for predicting either the persona of speaker 0 (Profile 0) or
of speaker 1 (Profile 1) given the dialogue utterances of speaker 0 (PERSON 0) or speaker
1 (PERSON 1). This is shown for dialogues between humans (PERSON 0) and either the 
KV Profile Memory model (``KV Profile'') which conditions on its own profile, or
the KV Memory model (``KV w/o Profile'') which does not.
  }
\end{table*}

\begin{table*}[t]
  \centering
  \begin{tabular}{ll|llllllll}
  \toprule
  \multirow{2}{*}{\textbf{Speaker}} & 
  \multirow{2}{*}{\textbf{Profile}} & 
  \multicolumn{8}{c}{\textbf{Dialogue Length}} 
  \\
   & & 1 & 2 & 3 & 4 & 5 & 6 & 7 & 8 \\
  \midrule
PERSON 0 & Profile 0  & 0.76 &  0.47 &  0.35 & 0.29 & 0.23 & 0.19 & 0.17  & 0.17 \\
PERSON 0 & Profile 1  & 0.51 &  0.39 &  0.32 & 0.29 & 0.27 & 0.27 &  0.25 & 0.25 \\  
PERSON 1 & Profile 0  & 0.57 &  0.52 &  0.48 & 0.46 & 0.45 & 0.43 &  0.43 & 0.43 \\
PERSON 1 & Profile 1  & 0.81 &  0.58 &  0.48 & 0.47 & 0.45 & 0.44 &  0.43 & 0.43 \\  
  \bottomrule
  \end{tabular}
  \caption{{\bf Profile Prediction By Dialog Length.}
  Error rates are given for predicting either the persona of speaker 0 (Profile 0) or
of speaker 1 (Profile 1) given the dialogue utterances of speaker 0 (PERSON 0) or speaker
1 (PERSON 1). This is shown for dialogues between humans (PERSON 0) and the 
KV Profile Memory model averaged over the first $N$ dialogue utterances from 100 conversations 
(where $N$ is the ``Dialogue Length''). The results show the  accuracy of predicting the persona 
improves in all cases as dialogue length increases.
     \label{tab:task2b}
  }
\end{table*}

\end{document}